\documentclass[journal]{IEEEtran}

\usepackage[utf8]{inputenc}
\usepackage[T1]{fontenc}
\usepackage{amsfonts,amsmath,amssymb}
\usepackage{anyfontsize}
\usepackage{booktabs}
\usepackage{caption}
\usepackage[inline]{enumitem}
\usepackage{fancyhdr}
\usepackage{graphicx}
\usepackage{mathtools}
\usepackage{microtype}
\usepackage{multirow}
\usepackage{placeins}
\usepackage{subcaption}
\usepackage{url}
\usepackage[table]{xcolor}
\usepackage{xspace}

\usepackage[pagebackref]{hyperref}
\hypersetup{%
	pdfauthor={Younes Belkada, Lorenzo Bertoni, Romain Caristan, Taylor Mordan, Alexandre Alahi},%
	pdftitle={Do pedestrians pay attention? Eye contact detection for autonomous driving},%
	colorlinks=true,%
	linkcolor=blue,%
	citecolor=green%
}
\usepackage[nameinlink]{cleveref}

\makeatletter
\g@addto@macro{\UrlBreaks}{\UrlOrds}
\makeatother

\DeclarePairedDelimiter\abs{\lvert}{\rvert}%
\DeclarePairedDelimiter\norm{\lVert}{\rVert}%

\makeatletter
\let\oldabs\abs
\def\abs{\@ifstar{\oldabs}{\oldabs*}}
\let\oldnorm\norm
\def\norm{\@ifstar{\oldnorm}{\oldnorm*}}
\makeatother

\newcommand{\eg}{e.g.\@\xspace}
\newcommand{\ie}{i.e.\@\xspace}
\newcommand{\etal}{et al.\@\xspace}

\title{Do Pedestrians Pay Attention?\\Eye Contact Detection in the Wild}

\author{Younes~Belkada$^{\star}$,
        Lorenzo~Bertoni$^{\star}$,
        Romain~Caristan,
        Taylor~Mordan,
        and~Alexandre~Alahi
\thanks{$^\star$ Equal contribution.}%
\thanks{Y. Belkada is with Sorbonne Université, France.}
\thanks{L. Bertoni, R. Caristan, T. Mordan and A. Alahi are with VITA, EPFL, Switzerland.}}

\begin{document}

\maketitle

\begin{abstract}
   In urban or crowded environments, humans rely on eye contact for fast and efficient communication with nearby people. Autonomous agents also need to detect eye contact to interact with pedestrians and safely navigate around them.
    In this paper, we focus on eye contact detection in the wild, \ie, real-world scenarios for autonomous vehicles with no control over the environment or the distance of pedestrians.
    
    We introduce a model that leverages semantic keypoints to detect eye contact and show that this high-level representation (i) achieves state-of-the-art results on the publicly-available dataset JAAD, and (ii) conveys better generalization properties than leveraging raw images in an end-to-end network. To study domain adaptation, we create LOOK: a large-scale dataset for eye contact detection in the wild, which focuses on diverse and unconstrained scenarios for real-world generalization. The source code and the LOOK dataset are publicly shared towards an open science mission.
\end{abstract}

\begin{IEEEkeywords}
    Autonomous Vehicles, Computer Vision, Deep Learning, Eye Contact Detection, Human Pose Estimation, Pedestrian Intention.
\end{IEEEkeywords}

\section{Introduction}

\IEEEPARstart{W}{hen} walking or driving, people use eye contact to communicate intentions, pay attention to their environments, or acknowledge the presence of others.
Autonomous agents also need to understand this implicit channel of communication to move naturally around humans and avoid collisions~\cite{rasouli2017agreeing, rasouli2019autonomous}.
Eye contact detection is especially useful for autonomous vehicles, as they need to understand whether a pedestrian intends to cross the street in front of the vehicle~\cite{rasouli2017are, varytimidis2018action}.
Similarly, smaller robots moving in crowds should be capable of detecting whether pedestrians have noticed them and are more likely to actively avoid them~\cite{kooij2014context, kothari2021human}. Finally, even in smart cities, eye contact detection can be useful to better understand pedestrians' behaviors, \eg, identify where their attentions go or what public signs they are looking at.

\begin{figure}[t]
    \centering
    \includegraphics[width=\linewidth]{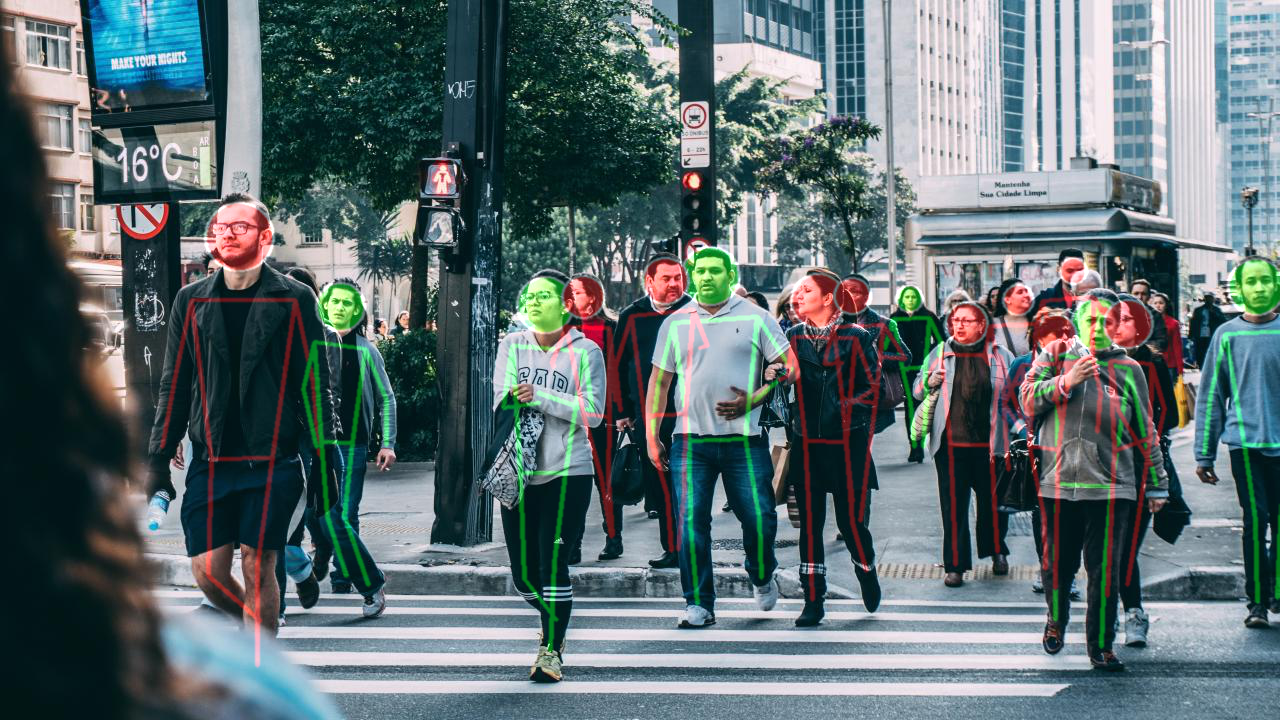}
    \caption{Typical scene for eye contact detection \textit{in the wild}, where pedestrians might be far from the camera and heavily occluded. Our method estimates, from predicted body poses, whether people are paying attention (showed in green) to the ego camera through eye contact, or are distracted (showed in red). This information can then help to better forecast their behaviors and to reduce the risk of collision with a self-driving agent.\protect\footnotemark}
    \label{fig:pull}
\end{figure}

Although humans make eye contact with each other at all times, detecting this action in the wild, \ie, with no constraint on the environment such as exemplified in Figure \ref{fig:pull}, presents a few challenges.
First, the action can be quick and subtle, happening with small head movements lasting as short as a few milliseconds.
Because of this small window, both spatially and temporally, the detection is hard, and can easily be affected by environmental conditions, such as lighting or distances of pedestrians.
Furthermore, eye contact has received little attention and few datasets have been annotated with it~\cite{rasouli2017are, rasouli2019pie}, when compared to more popular vision tasks such as object detection~\cite{pascalvoc} or 2D pose estimation~\cite{Lin2014MicrosoftCC}.
All these reasons make it more difficult for autonomous systems to detect eye contact effectively and to generalize to new environments.

\footnotetext{Image under license CC-0, \url{https://jooinn.com/images1280_/people-walking-on-pedestrian-lane-during-daytime.jpg}.}
In order to mitigate these issues, we propose to detect eye contact from high-level semantic keypoints, as displayed in Figure \ref{fig:pull}.
Although one could expect images to be a key input representation for eye contact detection, we show that we can use keypoints as input to escape the image domain, and process them with a simple, yet effective neural architecture.
For this, we first rely on a pose estimation step, which extracts semantic keypoints for all pedestrians in an image, using the off-the-shelf pose detector OpenPifPaf~\cite{kreiss2019pifpaf}.
Using pose features as input rather than images presents several advantages.
As pose needs less resolution than gaze, while also being annotated on more diverse datasets, it should be less affected by noise from environmental conditions, and predictions should generalize better to different scenarios and environments.
Poses are also much less dimensional than images, and do not require as much network capacity to be processed properly.
This allows the use of lighter networks, which should help prevent overfitting on the few scenarios annotated.
Finally, by leveraging these high-level features, we remove background information and reduce effects from changes in image statistics, allowing our model to focus solely on eye contact detection.

Since there are not many datasets covering a large variety of scenarios for eye contact, and these usually include a limited number of pedestrians~\cite{rasouli2017are, rasouli2019pie}, we argue that if a model is to be trained on them and deployed in the real world, it then must be particularly able to generalize well to new, uncontrolled conditions.
We suggest evaluating this through cross-dataset generalization, and we annotate three common autonomous driving datasets with this new task, namely KITTI~\cite{Geiger2013Kitti}, nuScenes~\cite{nuscenes}, and JRDB~\cite{martin2021jrdb}, to diversify the scenarios involving eye contact.
When evaluating our models, we show that using semantic keypoints leads to models generalizing better to various datasets and scenarios. 
We publicly release the annotations as a new dataset, which we refer to as LOOK\footnote{Dataset: \url{https://looking-vita-epfl.github.io}}, as well as the source code\footnotemark, towards an open science mission.
\footnotetext{Source code: \url{https://github.com/vita-epfl/looking}}

To summarize, our contributions are as follows:
\begin{itemize}
    \item We propose a deep learning model leveraging semantic keypoints, specially adapted to the challenges of eye contact detection;
    \item We publicly release LOOK, a diverse, large-scale dataset for eye contact detection in the wild, with numerous unique pedestrians and a focus on generalization across domains and scenarios, by annotating three common autonomous driving datasets;
    \item We suggest an evaluation protocol for eye contact with real-world generalization in mind, and show that our approach yields state-of-the-art results and strong generalization compared to image-based methods.
\end{itemize}

\section{Related Work}

\subsection{General eye contact}

Gaze estimation has received a lot of attention from the Computer Vision community, as a simpler alternative to eye tracking.
As for most other tasks, all leading approaches now rely on Deep Learning to get state-of-the-art results on the various benchmarks~\cite{Zhang2020ETHXGaze}.
In this paper, we focus on eye contact detection, which can be considered as a special case of gaze estimation.
There are multiple works that tackle this problem, \eg, Smith et al.~\cite{CAVE_0324} focus on gaze locking from eyes' visual appearances by masking out their surroundings, Park et al.~\cite{Park_2018_ECCV} transform single eye images into simplified pictorial representations to regress the angle of the gaze.
Some others focus also on real-time inference.
Fischer et al.~\cite{fischer2018rt} use a cascade of networks to localize heads and face landmarks, to align them to a predefined normalized face image for extracting eye patches, then compute gaze with a deep network.
Rowntree et al.~\cite{8945919} also use two networks for head detection and gaze estimation in order to speed up the overall pipeline.

The major issue with these works is that the benchmarks and the methods are not applied to \textit{in-the-wild} applications for autonomous vehicles, where the resolution of the pedestrian is low.
They usually focus on situations where people are rather close to the camera (\eg, inside a vehicle, in front of a computer), where the heads occupy larger regions in the images, and with simple or plain backgrounds, sometimes in controlled setups.
On the other hand, we focus on eye contact detection \textit{in the wild}, where there is no prior constraint on the type of environment pedestrians are in.

\subsection{Eye contact between pedestrians and vehicles}

From a driver's perspective, detecting eye contact is an important cue that indicates pedestrians' awareness of the traffic and future crossing intentions.
However, few datasets have annotated this action.
JAAD~\cite{rasouli2017are} and PIE~\cite{rasouli2019pie} are two such datasets, both focusing on pedestrians likely to cross the road in front of vehicles.
They therefore allow learning and evaluating eye contact directly from images.
Rasouli et al.~\cite{rasouli2017are} use an AlexNet  \cite{alexnet} to classify cropped images of pedestrians' heads but require bounding boxes to be given.
Varytimidis et al.~\cite{varytimidis2018action} have a similar approach where they use an SVM to classify features from a convolutional network applied to head crops, and then process their predictions with contextual information.
Mordan et al.~\cite{mordan2020detecting} jointly detect pedestrians and eye contact, along with other attributes, in a single network forward pass using multi-task fields.

In the context of pedestrian crossings, multiple works (in addition to the previous ones) use eye contact as an intermediate feature to better predict pedestrians' future behaviors.
Kooij et al.~\cite{kooij2014context} estimate head orientation as a cue for pedestrians' situational awareness, and use it with other indicators to predict their paths around the road with a Dynamic Bayesian Network.
Other approaches use a similar strategy for pedestrian awareness, \eg, Hariyono et al.~\cite{hariyono2016estimation} for estimating the risk of collision, Kwak et al.~\cite{kwak2017pedestrian} for prediction pedestrian intention at night time.

Eye contact detection is directly related to whether pedestrians pay attention to the incoming traffic.
In practice, detecting that they do not pay attention is as important.
One of the main reasons for that is the use of a phone that draws their attention away from the road.
Identifying phone-related activities is therefore an insightful cue to detect.
Rangesh et al.~\cite{rangesh2018vehicles} show the practical importance of having gaze annotations, both for eye contact between drivers and pedestrians, or phone-related distractions.
Going further to recognize actions implying a phone, Saenz et al.~\cite{saenz2021detecting} use a two-branch convolutional network to predict distracted behaviors due to phone usage from stereo image pairs. 

While detecting phone-related activities or pedestrian intentions are crucial tasks, eye-contact detection remains an essential channel of communication. Pedestrians may have the intention to cross but have they seen the upcoming car they should yield to? Contrarily, people may hold the phone but still pay attention to the upcoming traffic.

\section{LOOK Dataset}
We argue that eye contact detection is a crucial yet under-explored task
to enable autonomous agents to safely navigate around pedestrians. To promote research in this area, we show that the current datasets are not sufficiently diverse for data-driven methods, and we create a new large-scale dataset for eye contact detection in the wild.

\begin{table}
    \centering
    \caption{Dataset statistics. \textit{Frames} is the total number of frames in the datasets. \textit{Pedestrians} indicates the number of unique pedestrians, while \textit{Instances} counts the number of occurrences of pedestrians in all frames. In brackets, we mention the percentage of instances that are looking at the camera. JAAD~\cite{rasouli2017jaad} and PIE~\cite{rasouli2019pie} datasets include a very large number of instances but in comparison a very small number of different pedestrians. In contrast, our LOOK dataset includes in total 7,944 unique pedestrians from three continents, enabling exhaustive studies on cross-dataset generalization.}
    \label{tab:datasets}
    \setlength{\tabcolsep}{4pt}  
    \begin{tabular}{lccc}
        \toprule
        Dataset & Frames & Instances [\% looking] & Pedestrians \\
        \midrule
        JAAD \cite{rasouli2017jaad} &  82K & 133K [18\%] & 686 \\
        \rowcolor{gray!15}
        PIE \cite{rasouli2019pie} & 909K & 739K [9\%] & 1,842 \\
        \midrule
        Our LOOK-KITTI \cite{Geiger2013Kitti} & 1,391 & 4.630 [17\%] & 425 \\
        \rowcolor{gray!15}
        Our LOOK-JRDB \cite{martin2021jrdb} & 9,441 & 39K [18\%] & 399 \\
        Our LOOK-nuScenes \cite{nuscenes} & 2,216 & 13K [9\%] & 7,100 \\
        \midrule
        \rowcolor{gray!15}
        Our LOOK & 13K & 57K [16\%] & \textbf{7,944} \\
        \bottomrule
    \end{tabular}
\end{table}

\begin{figure*}[tbhp]
    \centering
    \includegraphics[width=0.8\linewidth]{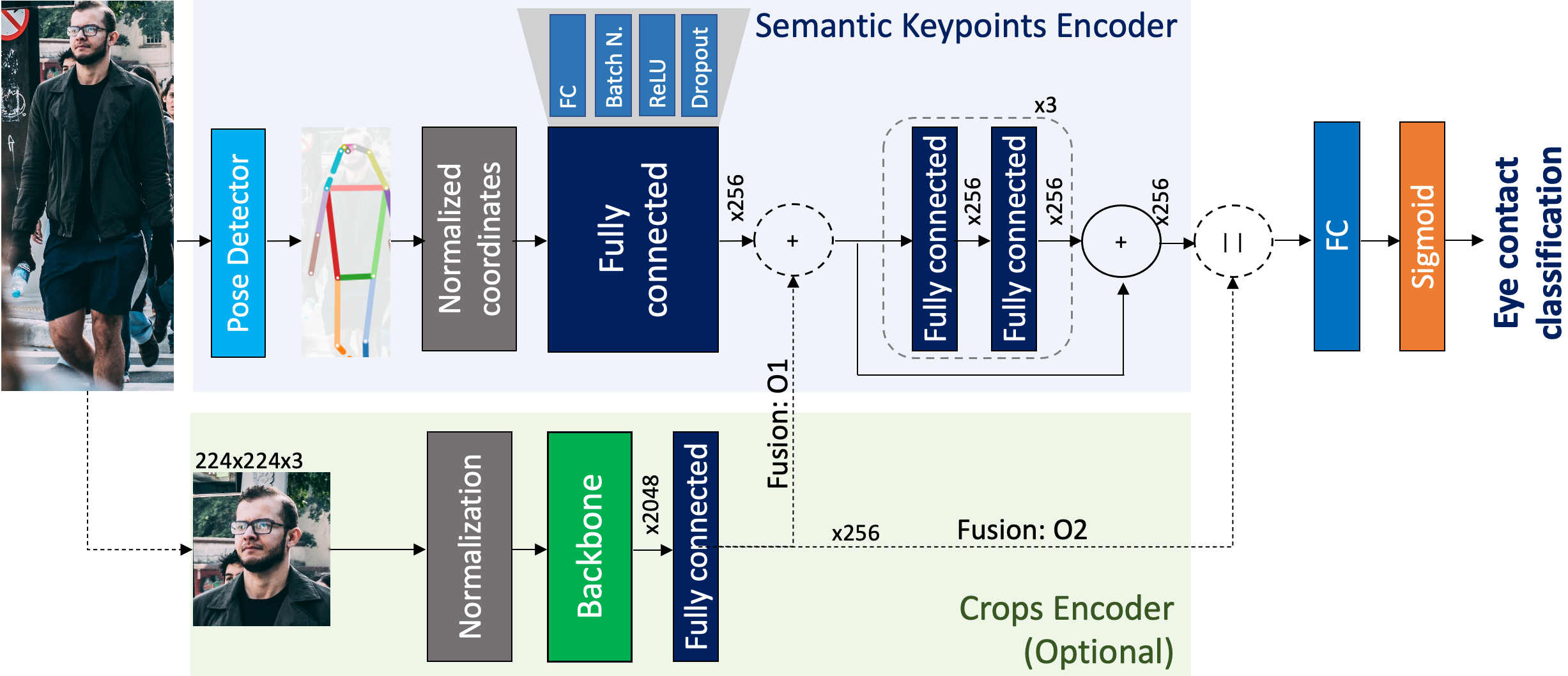}
    \caption{Modular architecture: the input of our keypoint-based model is the set of 2D joints extracted from a raw image, and the output is the binary flag indicating whether a person is looking at the camera. A \textit{Fully connected} block outputs 256 features and includes a fully connected layer (FC), a Batch Normalization layer (BN)~\cite{ioffe2015batch}, a ReLU activation function, and dropout~\cite{srivastava2014dropout}. Optionally, the features obtained from the semantic keypoints are concatenated with the features obtained from the head crops. We experiment with two types of fusions in the early (O1) or late (O2) layers, and with different convolutional architectures, such as ResNet-18 \cite{he2016residual} or ResNeXt-50 \cite{xie2017aggregated} as backbones for the crop-based module.}
    \label{fig:method}
\end{figure*}

\subsection{Existing datasets}
To the best of our knowledge, only two datasets contain annotations for the eye contact detection task: JAAD dataset \cite{rasouli2017jaad}, and PIE dataset \cite{rasouli2019pie}. 
JAAD consists of 390K instances of pedestrians labeled with bounding boxes and behaviour annotation, of which 17K instances have been labeled as looking at the driver (\ie, at the camera in the car). The dataset is large in size but limited in diversity. It is made of 346 video clips of 5-10 seconds recorded with an on-board camera at 30fps in North America and Europe. Thus, many frames show the same people in similar scenes, and the number of unique pedestrians looking at the camera is 686.

PIE \cite{rasouli2019pie} is also a recent dataset for pedestrian intention estimation. It shares many of the characteristics of its predecessor JAAD \cite{rasouli2017jaad}. It is recorded using an on-board camera at 30fps and consists of continuous footage of 6 hours in downtown Toronto, Canada. Out of 700K annotated pedestrian instances, there are 1,842 unique pedestrians, of which less than 180 are looking at the camera.

\subsection{Benchmark selection }
We have built a new large-scale dataset for eye contact detection in the wild by selecting publicly available images from three existing datasets: KITTI \cite{Geiger2013Kitti}, nuScenes \cite{nuscenes} and JRDB \cite{martin2021jrdb}. The first two are autonomous driving datasets and are made of images taken from a driver perspective. The latter one consists of videos taken from a small robot moving in crowded spaces inside Stanford University campus. In total, we have labeled 13,048 images from four different cities (Boston, Singapore, T\"ubingen, Palo Alto) in three continents. We aim for diversity, selecting pedestrians areas \cite{Geiger2013Kitti}, crowded images from six cameras around the car \cite{nuscenes}, and indoor environments from a robot perspective \cite{martin2021jrdb}. In total we have labeled around 8,000 unique pedestrians, making it the most diverse dataset for eye contact detection in the wild. Examples from the LOOK dataset are shown in Figures \ref{fig:ex-kitti}, \ref{fig:ex-nuscenes} and \ref{fig:ex-jrdb}.

We provide, together with the dataset annotation, the training and testing splits. We make sure that splits do not contain overlapping scenes and that the test set is sufficiently diverse, including 22\% of the total number of unique pedestrians over 15\% of the frames. 

\subsection{Annotation pipeline}
Our LOOK dataset has been annotated using the Amazon Mechanical Turk (AMT) platform. Each image has been annotated by four workers, which had the options to select whether a person was looking at the camera, somewhere else, or none of the two in case of ambiguity. We then include in the dataset only the labels with a consensus of at least three out of four annotators. This threshold has been selected by reviewing edge cases where not all the workers agree on a selected instance. 

To promote the creation of an ever-growing open-source dataset, we have also developed and released a labeling tool\footnotemark[\value{footnote}] to ease the annotation process. The tool leverages the off-the-shelf pose detector OpenPifPaf \cite{kreiss2019pifpaf} to locate the 2D bounding boxes of pedestrians. It then runs a pre-trained model (more detailed on Section \ref{sec:exp}) on the JAAD \cite{rasouli2017jaad} and PIE \cite{rasouli2019pie} dataset to provide a first guess. This pipeline allows annotators to only check and eventually correct wrong predictions.

To count the number of pedestrians, we use the tracking identification number for JRDB \cite{martin2021jrdb} dataset, while for the KITTI dataset \cite{Geiger2013Kitti}  we manually count them. In the case of the nuScenes dataset \cite{nuscenes}, we leverage the metadata associated with each frame. We approximate the number of unique pedestrians by only counting once the instances that appear in the same camera multiple times in a 5-seconds time window. We run sensibility analysis on the time window and opt for 5 second as the images are recorded from a moving vehicle in the majority of scenes.

\section{Eye Contact Detection}

The goal of our method is to detect from images whether humans are looking at the camera or somewhere else. We tackle autonomous driving scenarios, \ie, outdoor scenes where people may be several meters far from the camera. Our approach consists of two steps.
First, we use a 2D pose detector to obtain a low-dimensional representation from the image domain, which we call \textit{semantic keypoints}. The keypoints are a convenient representation that provides invariance to many factors, \eg, background artifacts, clothes, weather conditions. Second, we feed the extracted keypoints to a simple feed-forward neural network that detects the presence of eye contact.
In addition, we also explore multi-modal representations, by combining the keypoint representations with the features obtained from crops of images, and we explore different fusion techniques.
A diagram of our modular architecture can be found in Figure \ref{fig:method}.

\begin{table*}[t]
    \centering
    \caption{Comparing our proposed method and baseline results on JAAD \cite{rasouli2017jaad} and on our LOOK dataset. We evaluate eye contact classification using the average precision (AP) metric. For a fair comparison, we also report the recall of the detected pedestrians for each method. All approaches have been trained for classification on either JAAD solely, or on our LOOK dataset, and we evaluate them on both JAAD and LOOK. Our method is only trained on keypoints and reaches state-of-the-art results on the eye contact detection task on both the JAAD and LOOK datasets when compared with image- and crop-based methods. It also shows the best generalization properties when evaluated on a different dataset. The keypoints are obtained running an off-the-shelf pose estimator \cite{kreiss2019pifpaf} without re-training or adapting it to the different datasets.}
    \label{tab:main}
    \begin{tabular}{lllccccc}
        \toprule
        Training Dataset &
        Method & Input & 
        \multicolumn{5}{c}{Eye Contact Classification (AP) $\uparrow$ [Pedestrian Detection Recall $\uparrow$] }  \\
         \cmidrule(lr){4-8}
        & & & JAAD \cite{rasouli2017jaad}
         & LOOK-KITTI \cite{Geiger2013Kitti}
         & LOOK-JRDB \cite{martin2021jrdb}
         & LOOK-nuScenes \cite{nuscenes}
         & LOOK \\
        \midrule
        & Rasouli \cite{rasouli2017jaad} & Crops & 75.4 [80.1] & 65.9 [99.8] & 87.2 [98.2] & 78.7 [89.8] & 77.3 [95.9]\\
        \rowcolor{gray!15}
        & MTL-Fields \cite{mordan2020detecting} & Images & 82.6 [92.4] & 89.7 [93.1] & 82.1 [81.9] & 92.0 [71.8] & 87.9 [82.3] \\
        \multirow{-3}{*}{JAAD} & Our method & Keypoints & 85.9 [80.1] & 91.6 [99.8] & 94.8 [98.2] & 91.0 [89.8] & 92.5 [95.9]\\
        \midrule
        \rowcolor{gray!15}
        & Rasouli \cite{rasouli2017jaad} & Crops & 71.0 [80.1] & 76.8 [99.8] & 89.5 [98.2] & 82.9 [89.8] & 83.1 [95.9]\\
        & MTL-Fields \cite{mordan2020detecting} & Images &  80.7 [79.0] & 95.1 [96.5] & 95.2 [93.0] & 93.4 [68.4] & 94.6 [86.0]\\
        \rowcolor{gray!15}
        \multirow{-3}{*}{LOOK} & Our Method & Keypoints & \textbf{86.0} [80.1] & \textbf{96.4} [99.8] & \textbf{97.1} [98.2] & \textbf{95.1} [89.8] & \textbf{96.2} [95.9]\\
        \bottomrule
    \end{tabular}
\end{table*}

\subsection{Keypoint-based method}
We escape the image domain using 2D keypoints: a low-dimensional representation obtained through the off-the-shelf pose detector OpenPifPaf \cite{kreiss2019pifpaf,kreiss2021openpifpaf},  which was designed for crowded scenes and low-resolution images. The output of our network is the binary flag indicating whether a person is looking at the camera. 
To create the training and testing dataset, we match the bounding boxes enclosing the keypoints with the ground-truth bounding boxes using their intersection over union (IoU). We select the instances with the highest matching IoU above 0.3 for each ground truth. 

Keypoints are especially useful to prevent overfitting. To further increase generalization properties, we normalize the keypoints and zero-center them on the y-axis. Normalization prevents different scale differences from biasing the results, while the vertical location of a person in the image plane does not add any information regarding eye contact detection. The x-coordinate in the image plane, on the other side, may help infer the relative head and body orientations with respect to the camera.
For every keypoint $i$ with pixel coordinates $(u_i, v_i)$ in the image plane, we apply the following transformation:
\begin{equation}
\begin{cases}
    \hat{u_i} =& \frac{u_i - u_{hip}}{w_{box}} + \frac{u_{hip}}{w_{image}}, \\
    \hat{v_i} =& \frac{v_i - v_{hip}}{h_{box}},
\end{cases}
\end{equation}
where $(u_{hip}$, $v_{hip})$ is the mean of the coordinates of the left and right hips of the instance, $w_{box}$, $h_{box}$, the width and height of the enclosing box given by the keypoints, and $w_{image}$ the width of the input image. In practice, the normalization removes information on the size of the person as well as on the vertical location in the image plane.

Our architecture is composed of a simple fully-connected network with residual blocks \cite{he2016residual}, and includes batch-normalization \cite{ioffe2015batch} after every fully connected layer as well as dropout \cite{srivastava2014dropout}. The structure is inspired by the success in 3D vision tasks using 2D keypoints, especially Martinez \etal \cite{martinez2017simple} for 3D pose estimation, and Bertoni \etal \cite{monoloco} for human 3D localization. The residual blocks increase performances and avoid overfitting, while the model, which contains approximately 411K training parameters, is characterized by great speed and a low memory footprint. Its building blocks are shown in Figure~\ref{fig:method}.

\begin{table*}[t]
\setlength{\tabcolsep}{5pt}
    \centering
        \caption{Impact of different architectures on the AP metric for eye contact classification (\%) on different datasets. All methods have been trained on JAAD dataset \cite{rasouli2017jaad} only. \textit{\textbf{Crops}} stands for adapting a crop-based model first introduced by Rasouli \etal \cite{rasouli2017jaad} with a ResNet \cite{he2016residual} or ResNeXt \cite{xie2017aggregated} architecture. \textit{\textbf{Keypoints}} stands for our simple architecture only trained with keypoints as input, either all the 17 keypoints of the human body, or a subset of it: \textit{keypoints - Body} includes all the keypoints but the head ones, while \textit{Keypoints - Head} includes the ears, eyes and nose locations. \textit{\textbf{Keypoints \& Crops}} stands for our fusion-based approach combining keypoints and crops in a single representation. When training only using the 5 head keypoints, we obtain the best results on JAAD \cite{rasouli2017jaad} but training on all the keypoints generalizes better across datasets.}
    \label{tab:ablation}
    \begin{tabular}{lcccccc}
        \toprule
        Method
         & JAAD  \cite{rasouli2017jaad} & PIE \cite{rasouli2019pie} & LOOK-KITTI \cite{Geiger2013Kitti}  & LOOK-nuScenes \cite{nuscenes} & LOOK-JRDB \cite{martin2021jrdb} & LOOK \\
        \midrule
        \rowcolor{gray!15}
        Crops(ResNet-18 \cite{he2016residual} & 78.1 & 73.5 & 76.7 & 81.7 & 92.0 & 83.5 \\
        Crops (ResNeXt-50 \cite{xie2017aggregated}) & 79.7 & 74.2 & 72.0 & 85.7 & 92.5 & 83.4 \\
        \rowcolor{gray!15}
        Eyes Crops & 77.4 & 70.6 & 77.1 & 84.7 & 83.6 & 81.8 \\
        Keypoints & 85.9 & 83.8 & \textbf{91.6} & 91.0 & 94.8 & \textbf{92.5}  \\
        \rowcolor{gray!15}
        Body Keypoints & 76.4  & 72.6 & 79.3 & 80.7 & 75.4 & 78.5  \\
        Head Keypoints & \textbf{86.3} & \textbf{84.0} & 90.9 & 90.2 & \textbf{95.1} & 92.0  \\
        \rowcolor{gray!15}
        Keypoints \& Crops (ResNet-18 \cite{he2016residual}, Fusion: O1) & 78.0 & 75.2 & 79.7 & 85.3 & 91.6 & 85.5  \\
        Keypoints \& Crops (ResNet-18 \cite{he2016residual}, Fusion: O2)  & 78.7 & 75.6 & 78.9 & 84.3 & 92.7 & 85.4  \\
        \rowcolor{gray!15}
        Keypoints \& Crops (ResNeXt-50 \cite{xie2017aggregated}, Fusion: O1) & 79.5 & 75.1 & 73.6 & 85.8 & 92.1 & 83.8  \\
        Keypoints \& Crops (ResNeXt-50\cite{xie2017aggregated}, Fusion: O2)  & 80.6 & 75.9 & 74.1 & 86.2 & 93.2 & 84.5  \\
        \rowcolor{gray!15}
        Keypoints \& Eyes Crops (Fusion: O1)  & 83.9 & 79.9 & 87.0 & \textbf{91.2} & 92.5 &  90.2 \\
        \bottomrule
    \end{tabular}
\end{table*}

\subsection{Combined method}
We argue that 2D keypoints are a low-dimensional representation that contains enough information to understand whether a person is looking at the camera. This is motivated by the application we are targeting: autonomous driving scenarios, where people are often further away from the camera and the pupils may not be distinguishable. To test our hypothesis, we develop a modular architecture to optionally include visual information from the head region of a pedestrian.  We create a combined method that encodes features from both the keypoints and the cropped head region. While the former branch does not change, we select for the crops the upper third of the bounding box \cite{rasouli2017jaad} enclosing the keypoints, and we extract the features using a convolutional backbone. We explore different backbone architectures (\ie, AlexNet, \cite{alexnet} ResNet \cite{he2016residual} and ResNeXt \cite{xie2017aggregated}) and different fusion techniques. As visually described in Figure \ref{fig:method}, we experiment with \textit{early fusion} and with \textit{late fusion}. In the former option (O1), we sum the visual features extracted from a convolutional backbone with the raw features extracted from the 2D keypoints. In the latter option (O2), we concatenate the visual features together with the features extracted from the last layer of the fully-connected architecture. Our training schedule, inspired by Zamir \etal \cite{zamir2020robust}, consists of two steps. We first initialize the keypoint-based and the crop-based branches by training them independently. We then keep frozen all the layers before the concatenation and train the remaining ones. At both stages, we use the binary cross-entropy loss. The combined method allows us to verify whether adding visual information to the keypoint-based method increases the performance.

\section{Experiments} \label{sec:exp}

\subsection{Experimental setup}

\textbf{Evaluation metrics. }
We evaluate pedestrian detection and eye contact classification separately. Some previous methods \cite{rasouli2017jaad} do not include pedestrian detection, using a box classification approach, where the ground-truth boxes are given. In our case, a detection step is also included to ensure fair comparison among different methods. To disentangle the contributions of pedestrian detection and eye contact classification, we split the two tasks. To evaluate the detection results, we use the recall metric with a threshold on intersection over union (IoU) of 0.5. Compared to \cite{pascalvoc, mordan2020detecting}, we do not use Average Precision (AP) metric for detection as we only focus on instances labeled with the eye contact attribute, and in any given dataset very far instances are not annotated for it. In this case, the AP metric may penalize extra detections. In the classification setup, we evaluate the set of detected instances that match a ground-truth, and we use the AP metric to evaluate the classification of the binary attribute of looking or not at the camera. 

As shown in Table \ref{tab:datasets}, each dataset is unbalanced toward a majority of people not looking at the camera. 
Following again the procedure of \cite{mordan2020detecting}, we compute results on a balanced test set where negative instances are randomly sampled. The sampling is done 10 times to reduce the variance and the results are averaged.

 Regarding training and testing split, for JAAD dataset \cite{rasouli2017jaad} the official split is composed of 177 videos for training, 29 videos for validation, and 117 videos for testing. For PIE dataset
we use, as recommended, \textit{set01}, \textit{set02}, \textit{set04} for training, \textit{set05} \textit{set06} for validation and \textit{set03} for the testing set.

\textbf{Implementation details. }
To obtain input-output pairs of 2D joints and binary labels, we apply the off-the-shelf pose detector OpenPifPaf \cite{kreiss2019pifpaf, kreiss2021openpifpaf} and match our detections with the ground-truth boxes provided by each dataset. We train our keypoint-based architecture for 20 epochs, using binary cross-entropy loss function with Adam optimizer \cite{kingma2014adam}, with a learning rate of 0.0001, and mini-batches of 64 instances. The crop model is trained for 20 epochs, using binary cross-entropy loss function, SGD optimizer \cite{bottou2010large} with Nesterov momentum \cite{nesterov1983method}, a learning rate of 0.0001, and mini-batches of 32 instances. For the combined architecture, we pre-train both branches and freeze the early layers before the fusion of the features. We train the last layers with an SGD optimizer, a learning rate of 0.00001, and mini-batches of 128 instances.

The code, available online, is developed using PyTorch \cite{pytorch}. We do not apply any data augmentation procedure on the 2D poses.

\subsection{Baselines}
We argue that eye contact detection is a crucial task yet to be solved to develop safe autonomous vehicles. However, to the best of our knowledge, very few methods have reported results on the eye contact task in JAAD \cite{rasouli2017jaad} or PIE  \cite{rasouli2019pie}. Rasouli \etal \cite{rasouli2017jaad} proposed to use image crops of people as inputs to an AlexNet architecture \cite{krizhevsky2012imagenet} followed by fully connected layers. Their published results on the JAAD dataset \cite{rasouli2017jaad} used a smaller version of the dataset and randomly split the instances of the dataset. Hence, the same unique pedestrian in different time frames may appear both in training and testing sets. For a fair comparison, we have re-implemented this method and evaluated it on the recently released official JAAD split to prevent any contamination of the testing set.

In addition, we compare against the very recent MTL-Fields developed by Mordan \etal \cite{mordan2020detecting}.
It is a field-based approach that leverages multiple pedestrian attributes in a multi-task fashion, including eye contact, to understand the visual appearances and behaviors of pedestrians.
Contrary to Rasouli \etal \cite{rasouli2017jaad} that operate on image crops of people and therefore discard context around them, MTL-Fields keep full images in order to understand the scenes and learn interactions between pedestrians.
As their code is open-source, we train a network and evaluate it with our setup for eye contact detection.

\textbf{Our baselines. }
One of our goals is to compare the properties of keypoints and crops for the eye contact task. Thus, we develop a modular architecture that is either based on keypoints only, or can combine keypoints and visual information together, and we benchmark it with the following baselines:
\begin{itemize}
\item \textit{Crops}: when referring to methods using crops only, we consider the approach of Rasouli \etal \cite{rasouli2017jaad} with a more recent ResNet \cite{he2016residual} or ResNeXt  \cite{xie2017aggregated} backbone. Rasouli \etal \cite{rasouli2017jaad} train their model with ground-truth crops without including detection results. For a fair comparison, we train and evaluate the model including the same set of instances provided by the OpenPifPaf detector \cite{kreiss2021openpifpaf}.

\item \textit{Head \& Body Keypoints}: we test our keypoint-based architecture with subsets of keypoints, either only including the keypoints of the head region (eyes, nose, ears), or only the ones from the rest of the body. The goal is to analyze whether the body orientation also provides informative cues, or the head keypoints suffice for the eye contact detection task.

\item \textit{Keypoints \& Eye Crops}: we test whether adding visual information about the region around the eyes could be informative and less prone to overfitting than head crops. From the 2D keypoint locations of the eyes and ears, we crop a small region around the pupils and resize it to a fixed patch of 3x10x30 pixels. The model architecture is consistent with the one shown in Figure \ref{fig:method}, but we substitute the head crops with the eyes one, and a convolutional backbone with a fully connected block.
\end{itemize}

\begin{table}[t]
    \setlength{\tabcolsep}{4pt}
    \centering
    \caption{Evaluating cross-dataset results for our best crop- and keypoint-based methods using the AP binary classification metric (\%) on the JAAD dataset \cite{rasouli2017jaad}. In parenthesis, the relative difference with respect to the same method trained on the JAAD dataset \cite{rasouli2017jaad}. \textit{Instances} counts the total number of training instances.}
    \label{tab:cross-dataset}
    \begin{tabular}{lccc}
        \toprule
         & & \multicolumn{2}{c}{JAAD \cite{rasouli2017jaad}} \\
        \cmidrule(lr){3-4}
        Training Datasets & Instances
         & Crops & Keypoints  \\
         \midrule
        JAAD \cite{rasouli2017jaad} 
        & 50K & 79.7 ( - ) & 85.9 ( - )  \\
        \midrule
        LOOK-nuScenes \cite{nuscenes}
        & 10K & 71.1 (-8.6) & 84.6 (-1.3)  \\
        \rowcolor{gray!15}
        LOOK
        & 41K &  73.7 (-6.0) & 86.0 (+0.1)  \\
        LOOK + PIE \cite{rasouli2019pie}
        & 61K & 75.1 (-4.6) & \textbf{87.7} (\textbf{+1.8})   \\
        \bottomrule
    \end{tabular}
\end{table}

\subsection{Quantitative results} 
\begin{figure*}[t]
    \centering
    \begin{subfigure}[t]{\textwidth}
        \centering
        \includegraphics[width=\textwidth]{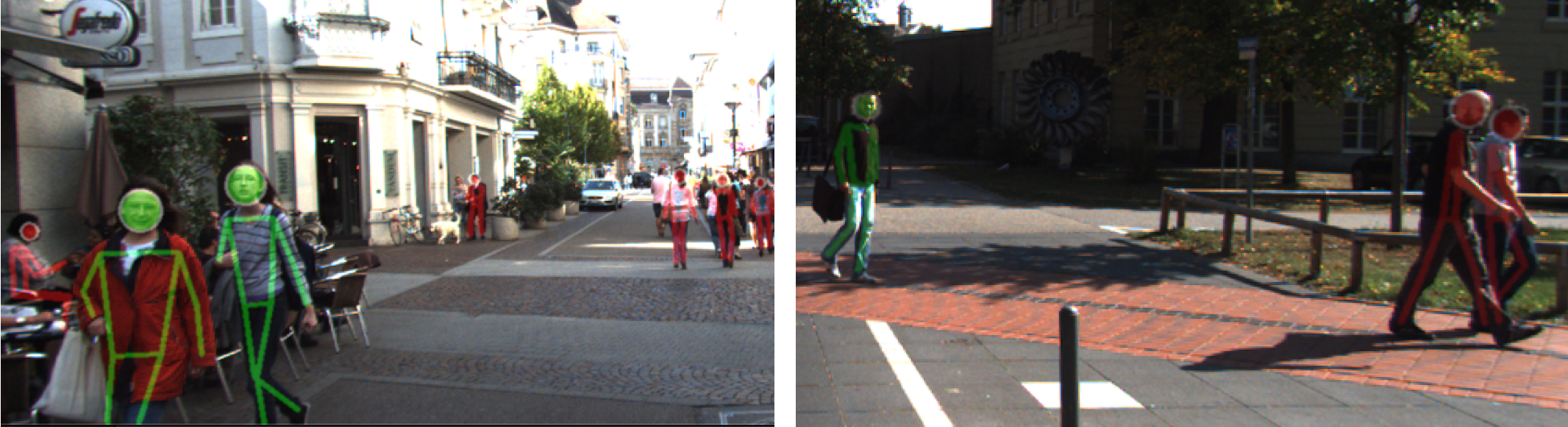}
        \caption{LOOK-KITTI~\cite{Geiger2013Kitti}}
        \label{fig:ex-kitti}
    \end{subfigure}
    \hfill
    \begin{subfigure}[t]{\textwidth}
        \centering
        \includegraphics[width=\textwidth]{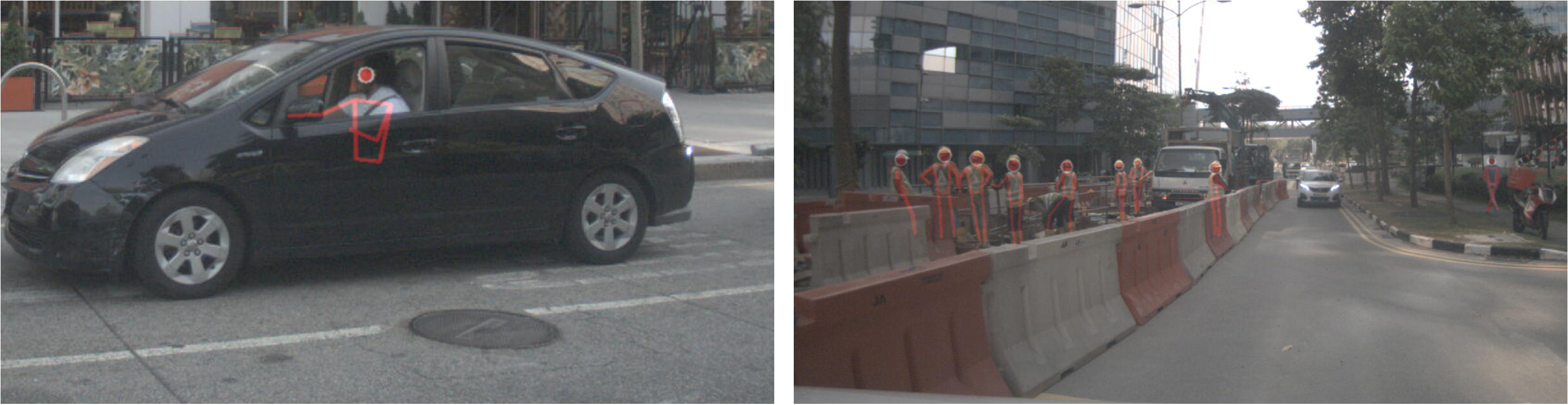}
        \caption{LOOK-nuScenes~\cite{nuscenes}}
        \label{fig:ex-nuscenes}
    \end{subfigure}
    \begin{subfigure}[t]{\textwidth}
        \centering
        \includegraphics[width=\textwidth]{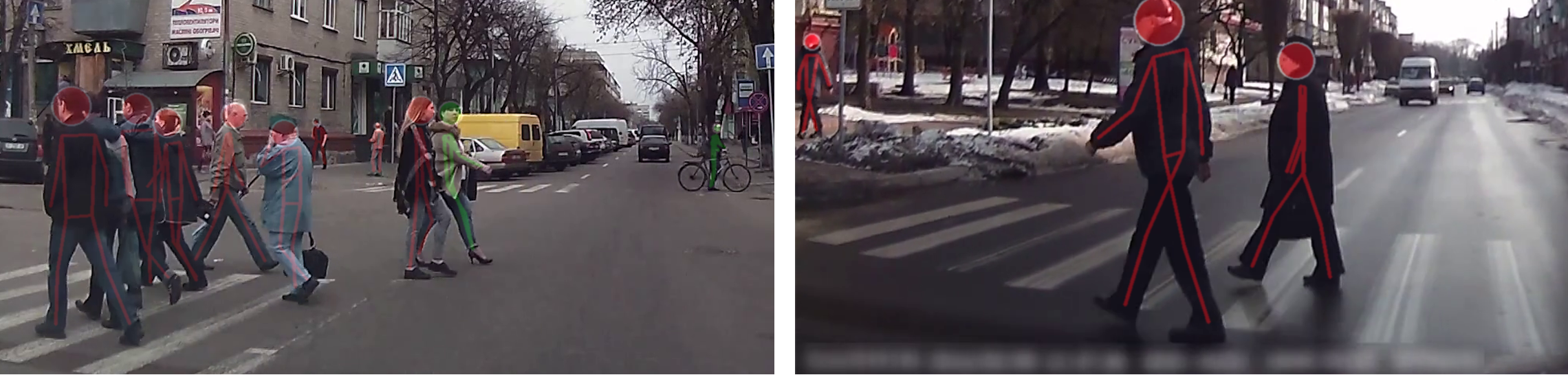}
        \caption{JAAD~\cite{rasouli2017jaad}}
        \label{fig:ex-jaad}

    \end{subfigure}
    \hfill
    \begin{subfigure}[t]{\textwidth}
        \centering
        \includegraphics[width=\textwidth]{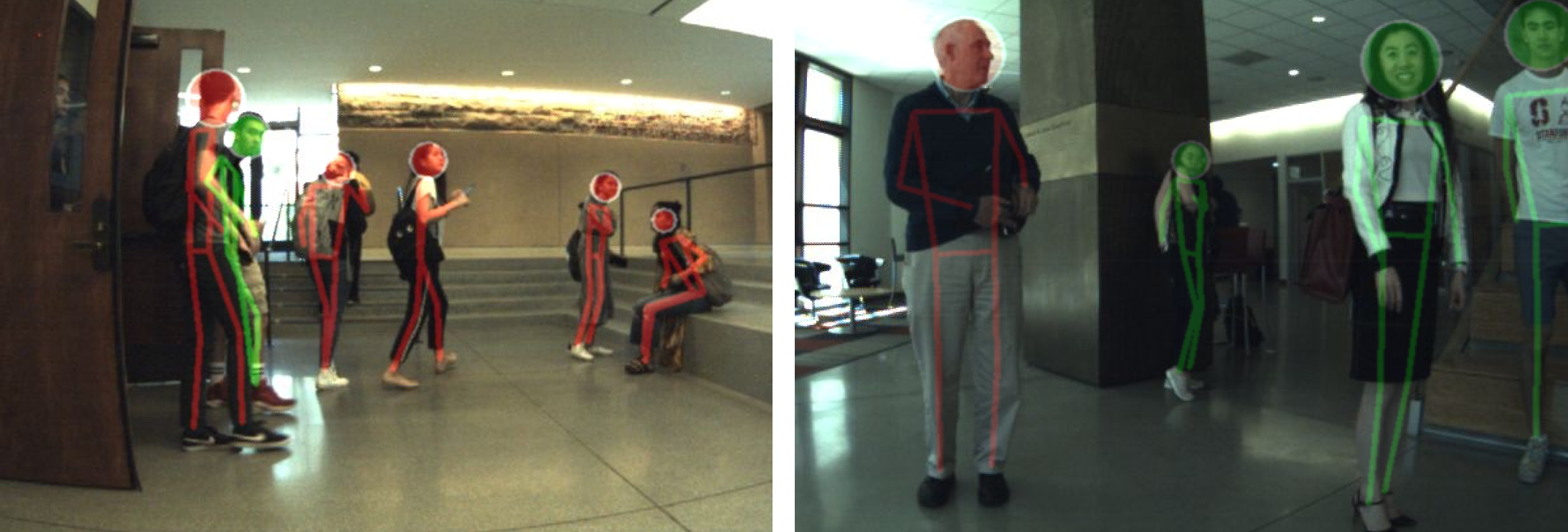}
        \caption{LOOK-JRDB~\cite{martin2021jrdb}}
        \label{fig:ex-jrdb}
    \end{subfigure}
    \caption{Qualitative results for the eye contact detection task on multiple datasets. People with green poses are predicted as looking at the camera, people with red poses as not looking.}
    \label{fig:ex}
\end{figure*}

In Table \ref{tab:main}, we show the results of training and evaluating each method on the same dataset (either JAAD \cite{rasouli2017jaad} or our LOOK dataset) as well as cross-dataset results. Our method achieves the best performances when compared to the other baselines, achieving at least a 5\% improvement both when testing on the same dataset and when evaluating cross-dataset generalization properties. More notably, our model is able to generalize well on our LOOK dataset when trained on the JAAD dataset \cite{rasouli2017jaad} only, as it reaches an AP of 92.5\%; almost on par when compared against baselines trained on the LOOK dataset. Qualitative examples from different datasets are shown in Figure \ref{fig:ex}.

Our recall results are shared with Rasouli \cite{rasouli2017jaad} baseline, as we train and evaluate their model on the same instances detected by the off-the-shelf pose detector OpenPifPaf \cite{kreiss2019pifpaf}. We observe that MTL-Fields \cite{mordan2020detecting} achieves higher recall on JAAD when trained on the same dataset, but the same recall drops by 15\% when trained on a different dataset. Our method on the other side maintains high recall when testing domain adaption, as it leverages an off-the-shelf pose estimator \cite{kreiss2019pifpaf} trained and optimized on a general-purpose dataset \cite{Lin2014MicrosoftCC} that is beneficial for domain adaptation.

\begin{figure}[b]
\centering
\includegraphics[width=\linewidth]{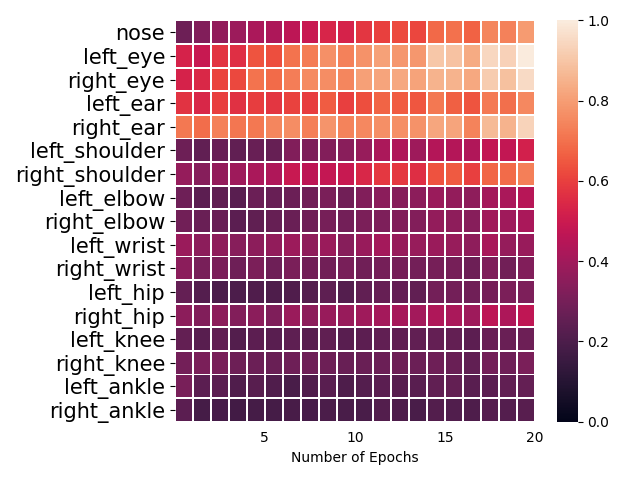}
\caption{Visual illustration of the normalized magnitude of the gradients of the loss function with respect to each keypoint during training. The keypoints related to the head (eyes and ears) are the ones that most affect the loss function. }
\label{fig:grads}
\end{figure}

\begin{table}[t]
    \centering
        \caption{Average precision (AP) in percentage (\%) as a function of the bounding box height in pixels for the JAAD dataset \cite{rasouli2017jaad}. Each cluster corresponds to one quartile of the distribution. For the crop-based methods, we consider our ResNeXt-50 model with late fusion when not differently specified.}
    \addtolength{\tabcolsep}{-0.8pt}
    \addtolength{\cmidrulekern}{-0.8pt}
    \begin{tabular}{lccccc}
        \toprule
         & \multicolumn{5}{c}{JAAD \cite{rasouli2017jaad}} \\
        \cmidrule(lr){2-6}
        Method / Box Height [px]
         & 240+ 
         & 160-240
         & 110-160
         & 0-110 
         & All \\
        \midrule
        \rowcolor{gray!15}
        Crops (ResNet-18) & 80.4 & 79.5 & 80.4 & 73.4 & 78.1 \\
        Crops (ResNeXt-50) & 79.0 & 81.2 & 83.0 & 75.3 & 79.7 \\
        \rowcolor{gray!15}
        Eyes Crops & 74.4 & 79.1 & 81.3 & 74.7 & 77.4 \\

        Keypoints & 87.9 & 88.7 & \textbf{87.0} & 78.7 & 85.9 \\
                \rowcolor{gray!15}
        Head Keypoints & \textbf{90.4} & \textbf{89.7} & 86.7 & 76.5 & \textbf{86.3} \\

        Keypoints \& Crops & 80.5 & 82.2 & 83.6 & 75.9 & 80.6 \\
        \rowcolor{gray!15}
        Keypoints \& Eyes Crops & 85.3 & 86.4 & 85.0 & \textbf{79.4} & 83.9 \\
        \bottomrule
    \end{tabular}
    \addtolength{\cmidrulekern}{0.8pt}
    \addtolength{\tabcolsep}{0.8pt}
    \label{tab:distance}
\end{table}

\subsection{Cross-dataset generalization. }
We further study cross-dataset generalization with our new LOOK dataset in Tables \ref{tab:ablation} and \ref{tab:cross-dataset}.  First of all, we investigate the performances of eleven crop- and keypoint-based methods in Table \ref{tab:ablation}. We train the models on JAAD dataset \cite{rasouli2017jaad} and evaluate them on JAAD \cite{rasouli2017jaad}, PIE \cite{rasouli2019pie}, and our LOOK dataset. Keypoint-based models perform and generalize better than crop-based ones, consistently with results obtained in Table \ref{tab:main}.  Surprisingly, combining visual information to the keypoints into our combined models (which we refer to as \textit{Keypoints \& Crops}) degrades the performances as it leads to stronger overfitting. The best results are achieved only by combining features from the eye region instead of the head region. We attribute this result to the generalization properties of keypoints, as this low-dimensional representation does not overfit on background scenes or specific faces. Yet even simpler models with no visual information achieve the best results on all the datasets, but for LOOK-nuScenes \cite{nuscenes}.

As additional experiment in Table \ref{tab:cross-dataset}, we train our best keypoint-based and crop-based methods on different datasets but JAAD~\cite{rasouli2017jaad}, and evaluate them on JAAD. We train our keypoint-based model on 10K instances from the nuScenes dataset~\cite{nuscenes}, and we obtain less than 2\% difference compared to  training it on the 50K instances of the JAAD dataset~\cite{rasouli2017jaad}. The best crop-based model, on the other side, achieves an AP of 71.1\%, down from an original 79.7\%. When increasing the number of instances from multiple datasets, the performances of the crop-based model never reach the baseline result of being trained on JAAd only. The keypoint-based model, on the other side, achieves the best performances on JAAD \cite{rasouli2017jaad} when trained on different datasets.

\subsection{Additional studies}
\textbf{The role of distance. }
We test the hypothesis that crop-based methods may be most effective when people are closer to the camera, while keypoints may be more useful when people are far away and details of the face are less informative. We obtain the distribution of bounding box heights for all the instances of the JAAD test set \cite{rasouli2017jaad}, and evaluate each quartile separately.  As we show in Table \ref{tab:distance}, the hypothesis is not verified: keypoints remain more effective than crops even for people close to the camera. This result may not be intuitive at first sight, but we are analyzing autonomous driving datasets, where even close people may be several meters away from the camera, and detecting the direction of the pupil may not be feasible. Keypoints provide a simple yet effective representation in these scenarios.

\textbf{Saliency map. }
To verify the impact of each keypoint in the final decision of the model, we compute the absolute value of the gradient of the objective function with respect to each input node for every epoch on the training set:

\begin{multline}
    i_k = \frac{1}{N} \sum_{j=0}^{N} \abs{\frac{\partial L(y_j, f(x_j))}{\partial k_{x}}} + \abs{\frac{\partial L(y_j, f(x_j))}{\partial k_{y}}} +  \\
    \abs{\frac{\partial L(y_j, f(x_j))}{\partial k_{c}}}  ,
\end{multline}
where $i_k$ represents the impact of the keypoint $k$ with its three components: $k_x$ and $k_y$ coordinates, and the confidence score $k_c$. We then average this value  by taking the mean absolute value across all the training instances $(x_j, y_j)$ that consists of $N$ samples. The results are illustrated in Figure \ref{fig:grads}. The dominant keypoints are the eyes and ears, as shown by the magnitude of the gradients of the loss function $L$ with respect to each keypoint. 

\begin{table}
    \centering
    \caption{Average run time performances for a single image on the JAAD test set. The detection steps for both Rasouli \cite{rasouli2017jaad} and our method are calculated with the off-the-shelf pose detector OpenPifPaf  \cite{kreiss2019pifpaf}, using either a ResNet-50 (R-50) \cite{he2016residual} or a ShuffleNetV2K30 (S-30) \cite{shufflenet} backbone.}

    \begin{tabular}{llcccc}
        \toprule
        & & \multicolumn{3}{c}{Average Run Time (ms)} \\
        \cmidrule(lr){3-5}
        Method & AP (\%) & Detection & Classification & Total \\
        \midrule
        Rasouli \cite{rasouli2017jaad} (S-30) & 71.0 & 602 & 39 & 672\\
        \rowcolor{gray!15}
        MTL-Fields \cite{mordan2020detecting} &  80.7 & -- & -- & 573 \\
        Our Method (S-30) & \textbf{85.9} & 602  & 0.8 & 626 \\
        \rowcolor{gray!15}
        Our Method (R-50) & 82.6 & \textbf{305} & 0.8 & \textbf{328} \\
        \bottomrule
    \end{tabular}
    \label{tab:time}
\end{table}

\textbf{Run Time. }
Our experiments have been conducted using a machine with a single NVIDIA GeForce GTX 1080 Ti and Intel(R) Core(TM) i7-8700 CPU @ 3.20GHz. In Table \ref{tab:time}, we compare the run time performances of several methods on the test set of the JAAD dataset \cite{rasouli2017jaad}. As the original method from Rasouli \etal did not include a detection step, we use the same backbone to extract the poses for our method and the crops for Rasouli's one. For MTL-Fields, detection and classification are performed in a single stage. Our method excels in the classification step with less than 1 ms of inference time as it uses low-dimensional keypoints. Regarding the detection step, our method is agnostic to the pose detector. We have tested it with OpenPifPaf \cite{kreiss2019pifpaf} using two different backbones and achieved the fastest run time with a ResNet-50 (R-50) \cite{he2016residual}.

\section{Conclusions}

Eye contact detection is a practically important task to better understand and forecast human behaviors.
In particular, autonomous robots need to solve this task to navigate safely around humans.
We have introduced a new deep learning approach for eye contact detection in the wild, \ie, with no prior knowledge on the environment, which is suited to the multiple challenges associated with this task.
We start by extracting semantic keypoints from images, and use them as low-dimension, high-level features to escape the image domain and focus on relevant information.
Then, we have compared several architectures to process this representation, including using it in addition to selected image crops.
We have also publicly released LOOK, a large-scale dataset for eye contact detection in the wild. We designed it with real-world generalization in mind, by annotating three common autonomous driving datasets to consider cross-dataset training and evaluation, and focus on multiple scenarios and diverse environments.
We evaluated our method and several approaches from the literature on LOOK to create a benchmark for this task, and show state-of-the-art results with robust generalization across datasets compared to image-based approaches.
We hope that this new benchmark can help foster further research from the community on this important but overlooked topic.

\bibliography{references}

\begin{thebibliography}{10}
\providecommand{\url}[1]{#1}
\csname url@samestyle\endcsname
\providecommand{\newblock}{\relax}
\providecommand{\bibinfo}[2]{#2}
\providecommand{\BIBentrySTDinterwordspacing}{\spaceskip=0pt\relax}
\providecommand{\BIBentryALTinterwordstretchfactor}{4}
\providecommand{\BIBentryALTinterwordspacing}{\spaceskip=\fontdimen2\font plus
\BIBentryALTinterwordstretchfactor\fontdimen3\font minus
  \fontdimen4\font\relax}
\providecommand{\BIBforeignlanguage}[2]{{%
\expandafter\ifx\csname l@#1\endcsname\relax
\typeout{** WARNING: IEEEtran.bst: No hyphenation pattern has been}%
\typeout{** loaded for the language `#1'. Using the pattern for}%
\typeout{** the default language instead.}%
\else
\language=\csname l@#1\endcsname
\fi
#2}}
\providecommand{\BIBdecl}{\relax}
\BIBdecl

\bibitem{rasouli2017agreeing}
A.~Rasouli, I.~Kotseruba, and J.~K. Tsotsos, ``Agreeing to cross: How drivers
  and pedestrians communicate,'' in \emph{IEEE Intelligent Vehicles Symposium
  (IV)}.\hskip 1em plus 0.5em minus 0.4em\relax IEEE, 2017, pp. 264--269.

\bibitem{rasouli2019autonomous}
A.~Rasouli and J.~K. Tsotsos, ``Autonomous vehicles that interact with
  pedestrians: A survey of theory and practice,'' \emph{IEEE Transactions on
  Intelligent Transportation Systems (T-ITS)}, vol.~21, no.~3, pp. 900--918,
  2019.

\bibitem{rasouli2017are}
A.~Rasouli, I.~Kotseruba, and J.~K. Tsotsos, ``Are they going to cross? a
  benchmark dataset and baseline for pedestrian crosswalk behavior,'' in
  \emph{Proceedings of the IEEE International Conference on Computer Vision
  (ICCV)}, 2017, pp. 206--213.

\bibitem{varytimidis2018action}
D.~Varytimidis, F.~Alonso-Fernandez, B.~Duran, and C.~Englund, ``Action and
  intention recognition of pedestrians in urban traffic,'' in \emph{14th
  International Conference on Signal-Image Technology \& Internet-based Systems
  (SITIS)}.\hskip 1em plus 0.5em minus 0.4em\relax IEEE, 2018, pp. 676--682.

\bibitem{kooij2014context}
J.~F.~P. Kooij, N.~Schneider, F.~Flohr, and D.~M. Gavrila, ``Context-based
  pedestrian path prediction,'' in \emph{Proceedings of the IEEE European
  Conference on Computer Vision (ECCV)}.\hskip 1em plus 0.5em minus 0.4em\relax
  Springer, 2014, pp. 618--633.

\bibitem{kothari2021human}
P.~Kothari, S.~Kreiss, and A.~Alahi, ``Human trajectory forecasting in crowds:
  A deep learning perspective,'' \emph{IEEE Transactions on Intelligent
  Transportation Systems (T-ITS)}, 2021.

\bibitem{rasouli2019pie}
A.~Rasouli, I.~Kotseruba, T.~Kunic, and J.~K. Tsotsos, ``{PIE}: A large-scale
  dataset and models for pedestrian intention estimation and trajectory
  prediction,'' in \emph{Proceedings of the IEEE International Conference on
  Computer Vision (ICCV)}, 2019, pp. 6262--6271.

\bibitem{pascalvoc}
M.~Everingham, S.~A. Eslami, L.~Van~Gool, C.~K. Williams, J.~Winn, and
  A.~Zisserman, ``The pascal visual object classes challenge: A
  retrospective,'' \emph{International journal of computer vision}, vol. 111,
  no.~1, pp. 98--136, 2015.

\bibitem{Lin2014MicrosoftCC}
T.-Y. Lin, M.~Maire, S.~J. Belongie, L.~D. Bourdev, R.~B. Girshick, J.~Hays,
  P.~Perona, D.~Ramanan, P.~Doll{\'a}r, and C.~L. Zitnick, ``Microsoft coco:
  Common objects in context,'' in \emph{The European Conference on Computer
  Vision (ECCV)}, 2014.

\bibitem{kreiss2019pifpaf}
S.~Kreiss, L.~Bertoni, and A.~Alahi, ``Pifpaf: Composite fields for human pose
  estimation,'' in \emph{The IEEE Conference on Computer Vision and Pattern
  Recognition (CVPR)}, 2019, pp. 11\,977--11\,986.

\bibitem{Geiger2013Kitti}
A.~Geiger, P.~Lenz, C.~Stiller, and R.~Urtasun, ``Vision meets robotics: The
  kitti dataset,'' \emph{International Journal of Robotics Research (IJRR)},
  2013.

\bibitem{nuscenes}
H.~Caesar, V.~Bankiti, A.~H. Lang, S.~Vora, V.~E. Liong, Q.~Xu, A.~Krishnan,
  Y.~Pan, G.~Baldan, and O.~Beijbom, ``nuscenes: A multimodal dataset for
  autonomous driving,'' \emph{arXiv preprint arXiv:1903.11027}, 2019.

\bibitem{martin2021jrdb}
R.~Martin-Martin*, M.~Patel*, H.~Rezatofighi*, A.~Shenoi, J.~Gwak, E.~Frankel,
  A.~Sadeghian, and S.~Savarese, ``{JRDB}: A dataset and benchmark of
  egocentric robot visual perception of humans in built environments,''
  \emph{IEEE Transactions on Pattern Analysis and Machine Intelligence
  ({TPAMI})}, 2021.

\bibitem{Zhang2020ETHXGaze}
X.~Zhang, S.~Park, T.~Beeler, D.~Bradley, S.~Tang, and O.~Hilliges,
  ``Eth-xgaze: A large scale dataset for gaze estimation under extreme head
  pose and gaze variation,'' in \emph{European Conference on Computer Vision
  (ECCV)}, 2020.

\bibitem{CAVE_0324}
B.~Smith, Q.~Yin, S.~Feiner, and S.~Nayar, ``{G}aze {L}ocking: {P}assive {E}ye
  {C}ontact {D}etection for {H}uman?{O}bject {I}nteraction,'' in \emph{ACM
  Symposium on User Interface Software and Technology (UIST)}, Oct 2013, pp.
  271--280.

\bibitem{Park_2018_ECCV}
S.~Park, A.~Spurr, and O.~Hilliges, ``Deep pictorial gaze estimation,'' in
  \emph{Proceedings of the European Conference on Computer Vision (ECCV)},
  September 2018.

\bibitem{fischer2018rt}
T.~Fischer, H.~J. Chang, and Y.~Demiris, ``{RT-GENE}: Real-time eye gaze
  estimation in natural environments,'' in \emph{Proceedings of the European
  Conference on Computer Vision (ECCV)}, 2018, pp. 334--352.

\bibitem{8945919}
T.~{Rowntree}, C.~{Pontecorvo}, and I.~{Reid}, ``Real-time human gaze
  estimation,'' in \emph{2019 Digital Image Computing: Techniques and
  Applications (DICTA)}, 2019, pp. 1--7.

\bibitem{alexnet}
\BIBentryALTinterwordspacing
A.~Krizhevsky, I.~Sutskever, and G.~E. Hinton, ``Imagenet classification with
  deep convolutional neural networks,'' in \emph{Advances in Neural Information
  Processing Systems}, F.~Pereira, C.~J.~C. Burges, L.~Bottou, and K.~Q.
  Weinberger, Eds., vol.~25.\hskip 1em plus 0.5em minus 0.4em\relax Curran
  Associates, Inc., 2012, pp. 1097--1105. [Online]. Available:
  \url{https://proceedings.neurips.cc/paper/2012/file/c399862d3b9d6b76c8436e924a68c45b-Paper.pdf}
\BIBentrySTDinterwordspacing

\bibitem{mordan2020detecting}
T.~Mordan, M.~Cord, P.~P{\'e}rez, and A.~Alahi, ``Detecting 32 pedestrian
  attributes for autonomous vehicles,'' \emph{arXiv preprint arXiv:2012.02647},
  2020.

\bibitem{hariyono2016estimation}
J.~Hariyono, A.~Shahbaz, L.~Kurnianggoro, and K.-H. Jo, ``Estimation of
  collision risk for improving driver's safety,'' in \emph{Conference of the
  IEEE Industrial Electronics Society (IECON)}.\hskip 1em plus 0.5em minus
  0.4em\relax IEEE, 2016, pp. 901--906.

\bibitem{kwak2017pedestrian}
J.-Y. Kwak, B.~C. Ko, and J.-Y. Nam, ``Pedestrian intention prediction based on
  dynamic fuzzy automata for vehicle driving at nighttime,'' \emph{Infrared
  Physics \& Technology}, vol.~81, pp. 41--51, 2017.

\bibitem{rangesh2018vehicles}
A.~Rangesh and M.~M. Trivedi, ``When vehicles see pedestrians with phones: A
  multicue framework for recognizing phone-based activities of pedestrians,''
  \emph{IEEE Transactions on Intelligent Vehicles}, vol.~3, no.~2, pp.
  218--227, 2018.

\bibitem{saenz2021detecting}
H.~Saenz, H.~Sun, L.~Wu, X.~Zhou, and H.~Yu, ``Detecting phone-related
  pedestrian distracted behaviours via a two-branch convolutional neural
  network,'' \emph{IET Intelligent Transport Systems}, vol.~15, no.~1, pp.
  147--158, 2021.

\bibitem{rasouli2017jaad}
A.~Rasouli, I.~Kotseruba, and J.~K. Tsotsos, ``Are they going to cross? a
  benchmark dataset and baseline for pedestrian crosswalk behavior,'' in
  \emph{Proceedings of the IEEE International Conference on Computer Vision
  Workshops}, 2017, pp. 206--213.

\bibitem{ioffe2015batch}
S.~Ioffe and C.~Szegedy, ``Batch normalization: Accelerating deep network
  training by reducing internal covariate shift,'' \emph{arXiv preprint
  arXiv:1502.03167}, 2015.

\bibitem{srivastava2014dropout}
N.~Srivastava, G.~Hinton, A.~Krizhevsky, I.~Sutskever, and R.~Salakhutdinov,
  ``Dropout: a simple way to prevent neural networks from overfitting,''
  \emph{The Journal of Machine Learning Research}, vol.~15, no.~1, pp.
  1929--1958, 2014.

\bibitem{he2016residual}
K.~He, X.~Zhang, S.~Ren, and J.~Sun, ``Deep residual learning for image
  recognition,'' in \emph{The IEEE Conference on Computer Vision and Pattern
  Recognition (CVPR)}, 2016, pp. 770--778.

\bibitem{xie2017aggregated}
S.~Xie, R.~Girshick, P.~Doll{\'a}r, Z.~Tu, and K.~He, ``Aggregated residual
  transformations for deep neural networks,'' in \emph{Proceedings of the IEEE
  conference on computer vision and pattern recognition}, 2017, pp. 1492--1500.

\bibitem{kreiss2021openpifpaf}
S.~Kreiss, L.~Bertoni, and A.~Alahi, ``{OpenPifPaf: Composite Fields for
  Semantic Keypoint Detection and Spatio-Temporal Association},'' \emph{arXiv
  preprint arXiv:2103.02440}, March 2021.

\bibitem{martinez2017simple}
J.~Martinez, R.~Hossain, J.~Romero, and J.~J. Little, ``A simple yet effective
  baseline for 3d human pose estimation,'' in \emph{The IEEE International
  Conference on Computer Vision (ICCV)}.\hskip 1em plus 0.5em minus 0.4em\relax
  IEEE, 2017, pp. 2659--2668.

\bibitem{monoloco}
L.~Bertoni, S.~Kreiss, and A.~Alahi, ``Monoloco: Monocular 3d pedestrian
  localization and uncertainty estimation,'' in \emph{The IEEE International
  Conference on Computer Vision (ICCV)}, October 2019.

\bibitem{zamir2020robust}
A.~R. Zamir, A.~Sax, N.~Cheerla, R.~Suri, Z.~Cao, J.~Malik, and L.~J. Guibas,
  ``Robust learning through cross-task consistency,'' in \emph{Proceedings of
  the IEEE/CVF Conference on Computer Vision and Pattern Recognition}, 2020,
  pp. 11\,197--11\,206.

\bibitem{kingma2014adam}
D.~P. Kingma and J.~Ba, ``Adam: A method for stochastic optimization,''
  \emph{arXiv preprint arXiv:1412.6980}, 2014.

\bibitem{bottou2010large}
L.~Bottou, ``Large-scale machine learning with stochastic gradient descent,''
  in \emph{Proceedings of COMPSTAT'2010}.\hskip 1em plus 0.5em minus
  0.4em\relax Springer, 2010, pp. 177--186.

\bibitem{nesterov1983method}
Y.~E. Nesterov, ``A method for solving the convex programming problem with
  convergence rate o (1/k\^{} 2),'' in \emph{Soviet Mathematics Doklady}, vol.
  269, 1983, pp. 543--547.

\bibitem{pytorch}
A.~Paszke, S.~Gross, F.~Massa, A.~Lerer, J.~Bradbury, G.~Chanan, T.~Killeen,
  Z.~Lin, N.~Gimelshein, L.~Antiga \emph{et~al.}, ``Pytorch: An imperative
  style, high-performance deep learning library,'' in \emph{Advances in Neural
  Information Processing Systems}, 2019, pp. 8024--8035.

\bibitem{krizhevsky2012imagenet}
A.~Krizhevsky, I.~Sutskever, and G.~E. Hinton, ``Imagenet classification with
  deep convolutional neural networks,'' \emph{Advances in neural information
  processing systems (NeurIPS)}, vol.~25, pp. 1097--1105, 2012.

\bibitem{shufflenet}
N.~Ma, X.~Zhang, H.-T. Zheng, and J.~Sun, ``Shufflenet v2: Practical guidelines
  for efficient cnn architecture design,'' in \emph{The European Conference on
  Computer Vision (ECCV)}, 2018, pp. 116--131.

\end{thebibliography}

\begin{IEEEbiography}[{\includegraphics[width=1in,height=1.25in,clip,keepaspectratio]{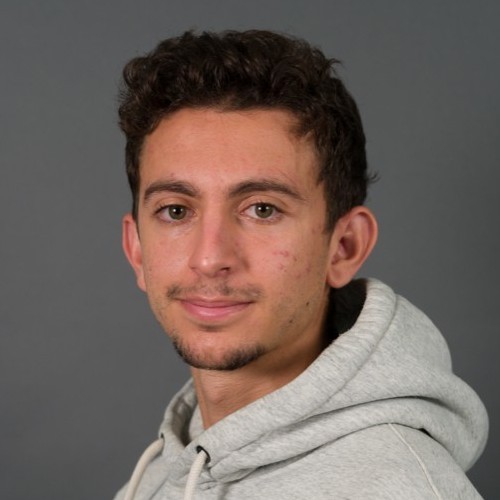}}]%
{Younes Belkada} has finished his Engineering Degree at Sorbonne University (Polytech Sorbonne), Paris, France. He is now enrolled in the MVA (Mathematics, Computer Vision, Machine Learning) Master's degree at ENS Paris Saclay.
He worked with VITA lab during his exchange semester at EPFL, Lausanne, Switzerland. During the first part of 2021, Younes was a Research Intern at Intel Corporation in Dublin, Ireland.
His interests focus on applying Deep Learning to real-world problems.
\end{IEEEbiography}

\begin{IEEEbiography}[{\includegraphics[width=1.05in,height=1.5in,clip,keepaspectratio]{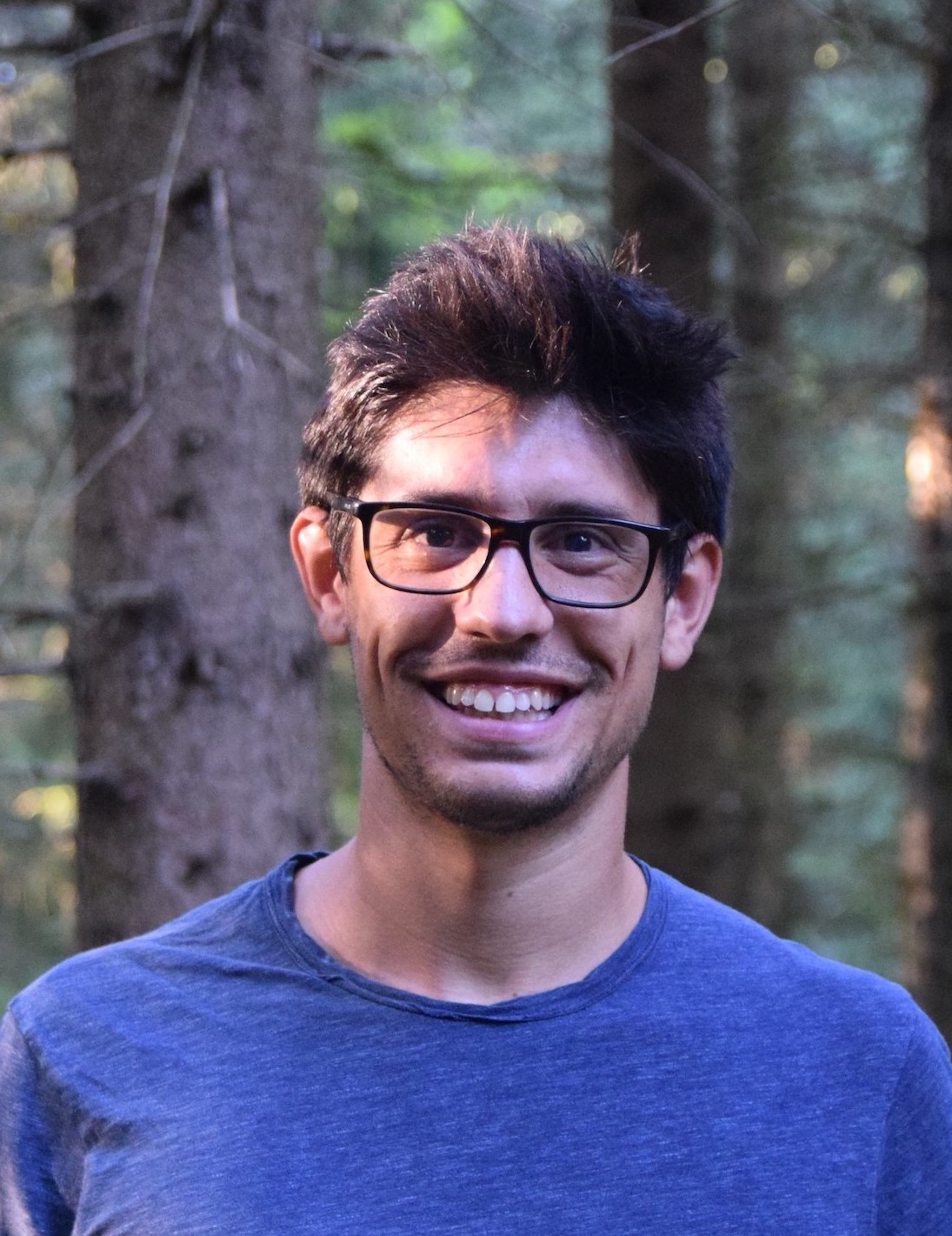}}]{Lorenzo Bertoni} received the master’s degree from the University of Illinois at Chicago and the master’s degree from the Polytechnic University of Turin. He is currently pursuing the Ph.D. degree with the Visual Intelligence for Transportation (VITA) Lab, EPFL, Switzerland, focusing on 3D vision for vulnerable road users. Before joining EPFL, he was a Management Consultant with Oliver Wyman and a Visiting Researcher with the University of California, Berkeley, working on predictive control for autonomous vehicles.
\end{IEEEbiography}

\begin{IEEEbiography}[{\includegraphics[width=1.05in,height=1.5in,clip,keepaspectratio]{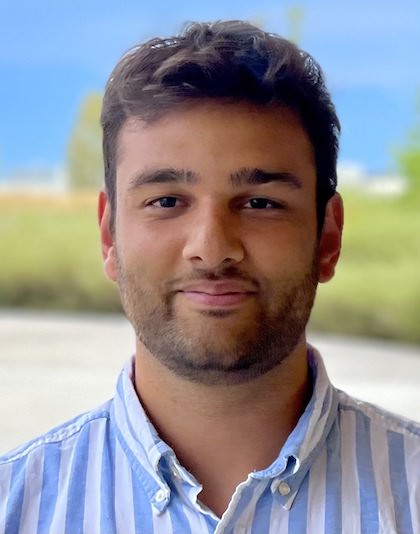}}]
{Romain Caristan} received his master's degree in Data Science at EPFL, Switzerland, and completed his master thesis in the VITA lab. His interests lie in using big data and machine learning techniques for autonomous driving applications.
\end{IEEEbiography}

\begin{IEEEbiography}[{\includegraphics[width=1in,height=1.25in,clip,keepaspectratio]{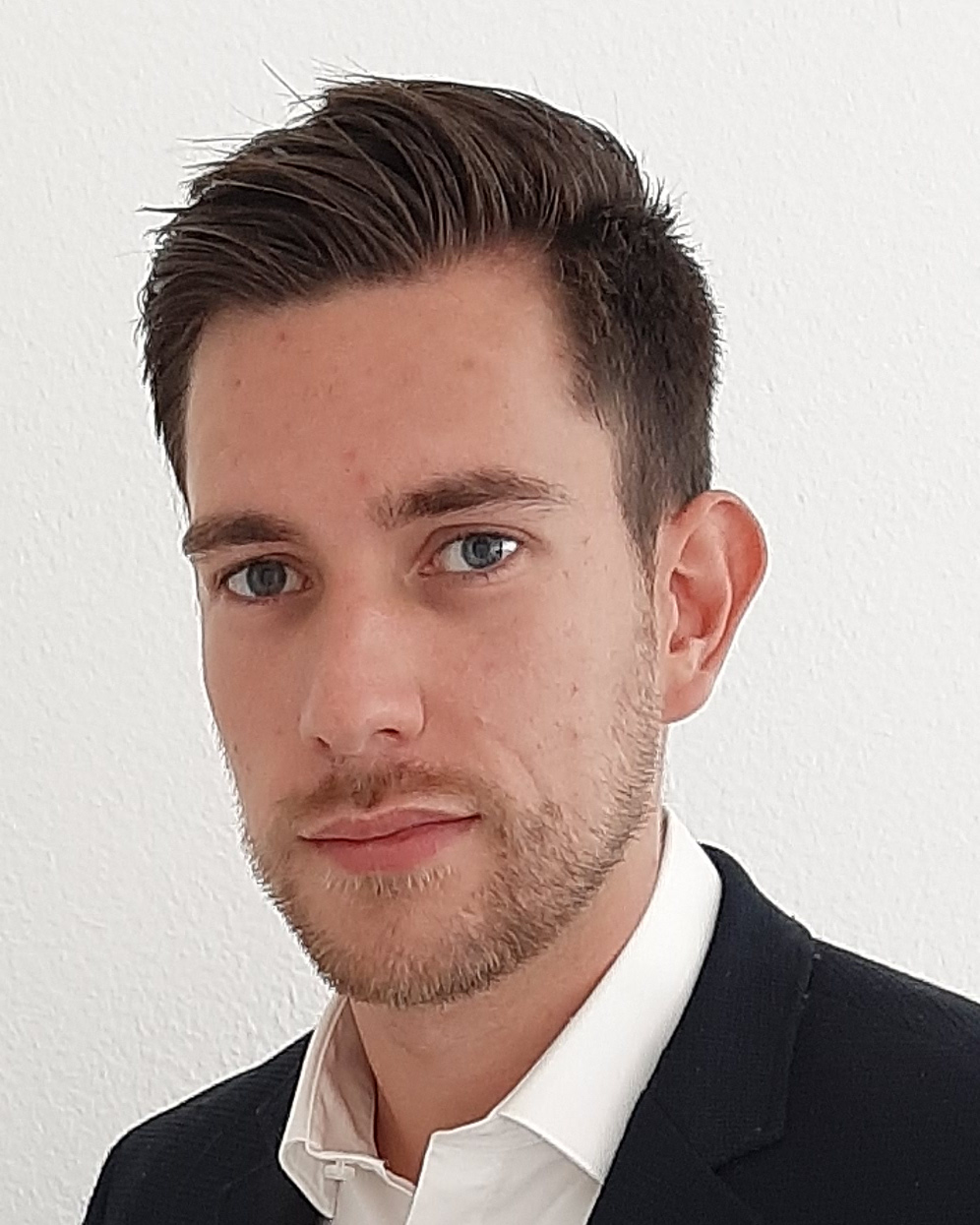}}]%
{Taylor Mordan}
received the engineering degree from ENSTA ParisTech, Paris, France, and the M.S. degree in computer science from UPMC, Paris, France, in 2015, then the Ph.D. degree in computer science from Sorbonne University, Paris, France, in 2018.
From 2015 to 2018, he was a Research Assistant with Thales LAS France. Since 2019, he has been a Post-Doctoral Researcher with VITA lab, EPFL, Lausanne, Switzerland. His research interests include computer vision, multi-task learning, and applications to perception in autonomous vehicles.
\end{IEEEbiography}

\begin{IEEEbiography}[{\includegraphics[width=1in,height=1.25in,clip,keepaspectratio]{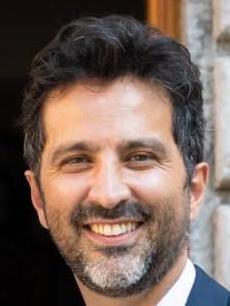}}]%
{Alexandre Alahi}
is currently an Assistant Professor at EPFL. He spent five years at Stanford University as a Post-doc and Research Scientist after obtaining his Ph.D. from EPFL. His research enables machines to perceive the world and make decisions in the context of transportation problems and smart environments. He has worked on the theoretical challenges and practical applications of socially-aware Artificial Intelligence, i.e., systems equipped with perception and social intelligence. 
\end{IEEEbiography}

\end{document}